\documentclass[fleqn,10pt]{wlscirep}
\usepackage[utf8]{inputenc}
\usepackage[T1]{fontenc}
\usepackage{lineno}
\usepackage{url}
\usepackage{hyperref}
\usepackage{comment}
\usepackage{xcolor}
\usepackage{siunitx}
\usepackage{longtable}
\usepackage{tabularx}
\usepackage{xspace}
\usepackage{float}
\usepackage{soul}
\usepackage{hhline}
\usepackage{csquotes}

\usepackage{hhline}
\usepackage{color, colortbl}
\definecolor{Gray}{gray}{0.9}
\usepackage{xcolor,bytefield}
\usepackage[colorinlistoftodos,prependcaption,textsize=tiny]{todonotes}

\usepackage[T1]{fontenc}

\title{FAIR principles for AI models with a practical application for accelerated high energy diffraction microscopy}

\author[1,2,3]{Nikil Ravi}
\author[1,2]{Pranshu Chaturvedi}
\author[1,4,*]{E. A. Huerta}
\author[1]{Zhengchun Liu}
\author[1]{Ryan Chard}
\author[1,5]{Aristana Scourtas}
\author[1,5]{K.J. Schmidt}
\author[1,4,5]{Kyle Chard}
\author[1,5]{Ben Blaiszik}
\author[1,4]{Ian Foster}

\affil[1]{Data Science and Learning Division, Argonne National Laboratory, Lemont, Illinois 60439, USA}
\affil[2]{Department of Computer Science, University of Illinois at Urbana-Champaign, Urbana, Illinois 61801, USA}
\affil[3]{Department of Mathematics, University of Illinois at Urbana-Champaign, Urbana, Illinois 61801, USA}
\affil[4]{Department of Computer Science, University of Chicago, Chicago, Illinois 60637, USA}
\affil[5]{Globus, University of Chicago, Chicago, Illinois 60637, USA}
\affil[*]{corresponding author: E.~A. Huerta, elihu@anl.gov}

\begin{abstract}
A concise and 
measurable set of FAIR (Findable, Accessible, Interoperable 
and Reusable) principles for scientific data is 
transforming the state-of-practice for data management 
and stewardship, supporting and enabling discovery 
and innovation. Learning from this initiative, and acknowledging the impact of 
artificial intelligence (AI) in the practice of 
science and engineering, we introduce 
a set of practical, concise, and measurable FAIR principles 
for AI models. We showcase how to create and share FAIR 
data and AI models within a unified computational 
framework combining the following elements: 
the 
Advanced Photon Source at 
Argonne National Laboratory, 
the Materials Data 
Facility, the Data 
and Learning Hub for Science, and 
funcX, 
and the Argonne Leadership Computing 
Facility (ALCF), in particular the ThetaGPU supercomputer 
and the SambaNova DataScale\textsuperscript{\textregistered} system  
at the ALCF AI Testbed.
We describe how this domain-agnostic computational framework 
may be harnessed to enable autonomous AI-driven discovery. 
\end{abstract}

\begin{document}

\flushbottom
\maketitle

\thispagestyle{empty}


\section*{Introduction}

\noindent Innovation at the interface of artificial intelligence (AI)  
and high performance computing is 
powering breakthroughs in science, engineering,
and industry~\cite{data2vec_pap,DeepLearning, hep_kyle,Nat_Rev_2019_Huerta,Narita2020ArtificialIP,gw_nat_ast,ai_agriculture,Uddin2019ArtificialIF,fair_hbb_dataset,Huerta:2021ybd,ai_math}. 
Thus, it is timely and important to define 
best AI practices that facilitate cross-pollination of expertise, 
reduce time-to-insight, increase reusability of scientific 
data and AI models by humans and machines, and 
reduce duplication of effort. To realize these goals, researchers 
are working to understand how to adapt FAIR 
guiding principles~\cite{fairguiding,fairmetrics}---originally 
developed in the context of digital assets, such as 
data and the tools, algorithms, and workflows that produce such 
data---to streamline 
the development and adoption of AI methodologies. 

FAIR guiding principles in the context of scientific datasets 
aim to accelerate innovation and scientific 
discovery by defining and implementing best practices for 
data stewardship and governance that enable the automation 
of data management. FAIR scientific datasets are also 
AI-ready when they are shared and published in suitable 
formats (\texttt{HDF5}~\cite{hdf5} or \texttt{ROOT}~\cite{root_cite}) 
that facilitate their use in modern computing environments 
and with open source APIs for AI 
research (\texttt{TensorFLow}~\cite{TensorFlow} 
or \texttt{PyTorch}~\cite{paszke2017automatic}).

The creation of FAIR and AI-ready datasets is transforming the 
state-of-practice of AI research across disciplines~\cite{fair_hbb_dataset,Sinaci2020FromRD,hpcfair,deagen_nat_sci_dat}. 
In view of these activities, and given the growth 
and impact of AI for Science programs, it is critical to 
define at a practical level what FAIR means for AI 
models---the theme of this article. To contextualize the 
FAIR principles for AI models that we introduce in this article, 
we begin by describing a FAIR 
and AI-ready scientific 
dataset that we used to create and publish FAIR AI models. 
We do this because there is an agreed-upon set of 
guidelines to FAIRify scientific datasets~\cite{fairguiding,fairmetrics,fair_hbb_dataset} that 
we take as the basis, or common ground, from which we 
define a set of practical FAIR principles for AI models. 

To quantify the FAIRness of 
our AI models, we have created a domain-agnostic computational 
framework that brings together advanced 
scientific data infrastructure and modern computing environments 
to conduct automated, accelerated, and reproducible 
AI inference. This computational framework facilitates the 
integration of 
FAIR and AI-ready datasets with FAIR AI models. 
This unified approach will enable researchers to obtain a clear 
understanding of 
the state-of-practice of 
AI to address contemporary scientific grand challenges. 
Understanding needs and gaps in available datasets and AI models 
will catalyze the 
sharing of knowledge and expertise at an accelerated pace and scale, 
thereby enabling focused R\&D in areas where AI capabilities 
are currently lacking. The components of this computational framework 
encompass: 

\begin{itemize}[nosep]
    \item Data 
and Learning Hub for Science (https://www.dlhub.org, DLHub)~\cite{dlhub, blaiszik_foster_2019} to 
publish and share FAIR AI models,
\item Materials Data 
Facility (https://materialsdatafacility.org, MDF)~\cite{mdf_article} to publish and share 
FAIR and AI-ready datasets,
\item funcX, https://funcx.org~\cite{chard2020funcx}, 
a distributed Function as a Service (FaaS) platform, 
to connect FAIR AI models hosted at DLHub 
and FAIR and AI-ready datasets hosted at  
MDF with the
ThetaGPU supercomputer at the Argonne Leadership Computing 
Facility (https://www.alcf.anl.gov, ALCF) to conduct reproducible AI-driven inference.
\end{itemize}

\noindent Below we describe how to use this domain-agnostic 
computational framework to evaluate the FAIRness of AI 
models. We also describe how to integrate uncertainty quantification 
metrics in AI models to assess the 
reliability of their predictions. Throughout our discussion, 
we leverage disparate hardware architectures, ranging 
from GPUs to the SambaNova Reconfigurable 
Dataflow Unit\textsuperscript{TM} (RDU) at 
the ALCF AI-Testbed, https://www.alcf.anl.gov/alcf-ai-testbed,  
to ensure that our 
ideas provide useful, easy-to-follow guidance to 
a diverse ecosystem of researchers and AI practitioners. 

\begin{figure}[htbp]
\centerline{
\includegraphics[width=\textwidth]{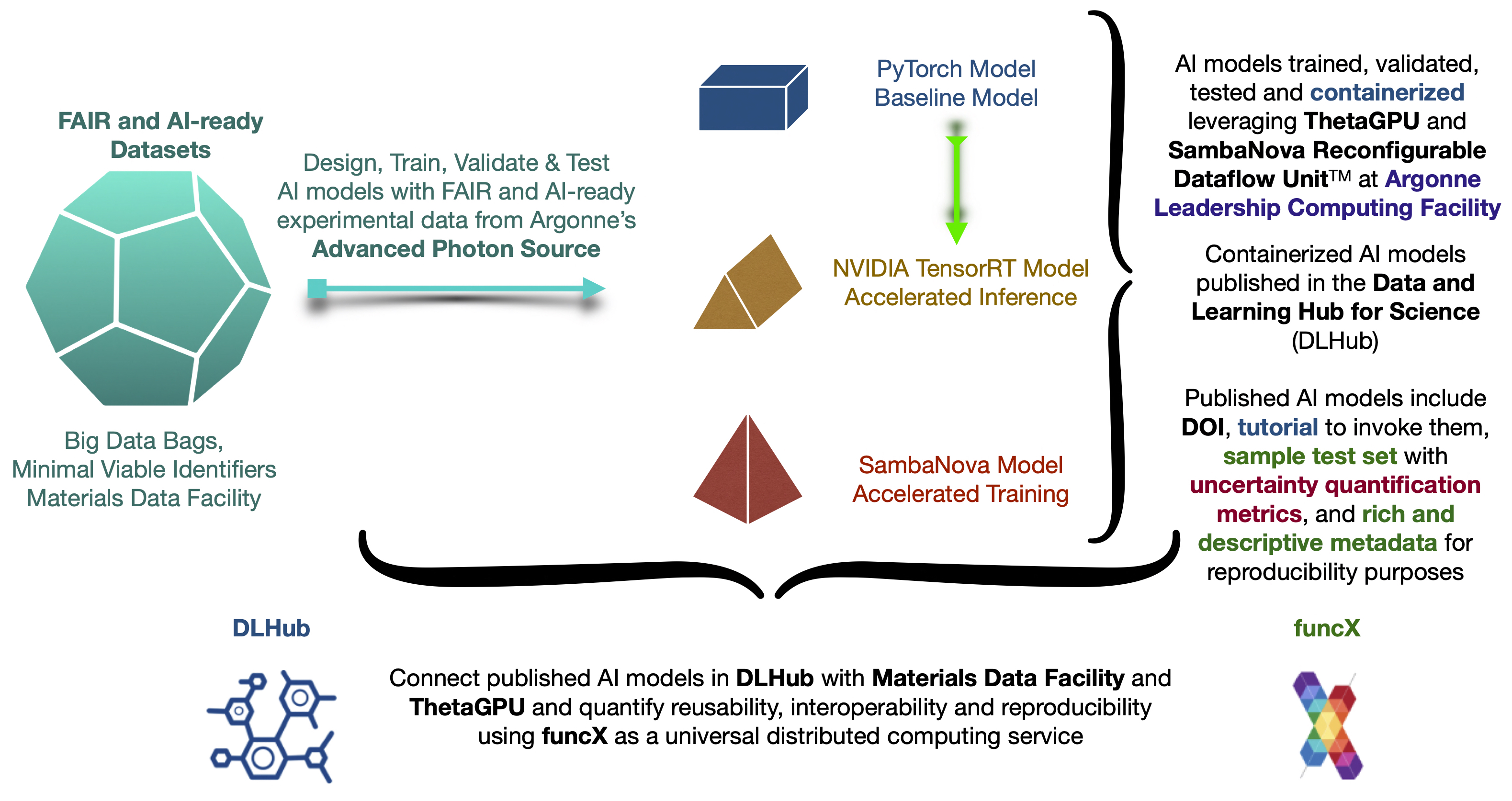}
}
\caption{\textbf{FAIR AI models and data} Proposed 
approach to combining FAIR and AI-ready experimental data, 
FAIR AI models and scientific data, and 
computing infrastructure to accelerate and 
automate scientific discovery.}
\label{fig:fair_ai}
\end{figure}

We release our FAIR AI datasets and models, as well as 
scientific software to enable 
other researchers to reproduce our work, and 
to engage in meaningful interactions that 
contribute towards an agreed upon, community wide  
definition of what FAIR means in the context of AI models, 
with an emphasis on practical applications. 
Figure~\ref{fig:fair_ai} summarizes 
our proposed approach to creating FAIR AI models 
and experimental datasets. This approach has three main 
pillars, namely: (i) the creation and sharing and 
FAIR and AI-ready 
datasets; (ii) the combination of such 
datasets with modern computing 
environments to streamline and automate the 
creation and publication of FAIR 
AI models; and (iii) the combination of FAIR datasets 
and AI models with scientific data infrastructure and advanced 
computing resources to automate accelerated AI inference with 
well-defined uncertainty quantification metrics to measure 
the reliability, reproducibility, 
and statistical robustness of AI predictions.

We selected high energy diffraction microscopy as a science 
driver for this work~\cite{BraggNN-IUCrJ}. This technique is used 
to characterize 
3D information about the structure of polycrystalline materials 
through the identification of Bragg diffraction peaks. The data 
used for the identification of Bragg peaks with our AI models 
was produced at the 
Advanced Photon Source, https://www.aps.anl.gov, at 
Argonne National Laboratory. While we guide the discussion 
of our methods with this application, the definitions, 
approaches and computational framework introduced in this 
article are domain-agnostic and may be harnessed for any 
other scientific application.

\section*{Results}

\noindent We present two main results:
\begin{itemize}[nosep]

    \item \textbf{FAIR and AI-ready experimental datasets} We 
    published experimental datasets used to train, 
    validate and test our AI models via 
    the MDF, 
    and created 
    Big Data Bags (\texttt{BDBags})~\cite{chard16togo} 
    with associated Minimal Viable Identifiers (\texttt{minids})~\cite{chard16togo} 
    to specify, describe, and reference each of these 
    datasets. We also published Jupyter notebooks via MDF
    that describe these datasets and illustrate how to 
    explore and use them to train, validate, and test AI 
    models. We FAIRified these datasets following  
    practical guidelines~\cite{fairguiding,fairmetrics,fair_hbb_dataset}.
    
    \item \textbf{Definition of FAIR for AI models and 
    practical examples} We use the aforementioned FAIR 
    and AI-ready datasets to produce three AI models: 
    a baseline \texttt{PyTorch} model, an 
    optimized AI model for accelerated inference that is 
    constructed by porting the baseline \texttt{PyTorch} model 
    into an NVIDIA \texttt{TensorRT} engine, 
    and a model created on the SambaNova DataScale\textsuperscript{\textregistered} system at the 
    ALCF AI-Testbed.
    We use the ThetaGPU supercomputer 
    at ALCF 
    to create and containerize the first two models. 
    We use these three models to 
    showcase how our proposed definitions for FAIR AI 
    models may be quantified by creating a framework 
    that brings together DLHub, 
    funcX, the MDF and disparate 
    hardware architectures at ALCF. 
\end{itemize}

\subsection*{FAIR and AI-ready Datasets} We FAIRified a 
high energy microscopy dataset produced at the 
Advanced Photon 
Source at Argonne National Laboratory. 
We split this dataset into a training dataset that we used 
to create the AI models 
described below, and 
a validation dataset that we used to compute relevant 
metrics pertaining to the performance of our AI models 
for regression analyses. We 
published these datasets via 
MDF and provide 
the following information:

\paragraph{Findable}
\begin{itemize}[nosep]
 \item Unique Digital Object Identifier (DOI) for Training\_Set~\cite{braggnn-training-set} andValidation\_Set~\cite{braggnn-validation-set}
 \item Rich and descriptive metadata
 \item Detailed description of the datasets, including data type and 
    shape
    \item Metadata uses machine-readable keywords
    \item Metadata contains resource identifier
    \item (Meta)data are indexed in a searchable resource
\end{itemize}

\paragraph{Accessible}
\begin{itemize}[nosep]
\item Datasets are published with CC-BY 4.0 licenses
\item Training\_Set~\cite{braggnn-training-set} and Validation\_Set~\cite{braggnn-validation-set} are open datasets
\item (Meta)data are retrievable by their identifier using a 
standardized communications protocol
\item Metadata remains discoverable, even in the absence 
of the datasets
\end{itemize}

\paragraph{Interoperable}
\begin{itemize}[nosep]
\item Datasets are published in open \texttt{HDF5} format 
\item Metadata contains qualified references to related data~\cite{zhengchun_repo_data} and publications~\cite{BraggNN-IUCrJ}
\item Metadata follows standards for X-ray spectroscopy, uses controlled vocabularies, ontologies and good data model following a physics classification 
scheme developed by the American Physical Society (APS)~\cite{aps_classification}. 
These datasets were collected and curated to 
streamline their use for AI analyses~\cite{sharma_data}
\item Uses FAIR vocabulary following the ten rules 
    provided in Ref.~\cite{10rules}
\end{itemize}

\paragraph{Reusable}
\begin{itemize}[nosep]
\item Author list, and points of contact including email addresses
\item Datasets include Jupyter notebooks to explore, understand 
and visualize the datasets
\item Datasets  
include Jupyter notebooks that show how to use the 
Training\_Set~\cite{braggnn-training-set} to 
train AI models with the scripts we have released in GitHub~\cite{zhengchun_repo}
\item Datasets  
include Jupyter notebooks that show how to use the 
Validation\_Set~\cite{braggnn-validation-set} to conduct AI inference using the computational 
framework we described above
\item Datasets include a description of how they were produced and 
provide this information in a machine-readable metadata format
\end{itemize}

\noindent We have also created \texttt{BDBags} and \texttt{minids} for each 
of these datasets to ensure that creation, 
assembly, consumption, identification, and exchange of 
these datasets can be easily integrated into a user's 
workflow. A \texttt{BDBag} 
is a mechanism for defining a dataset and its contents by 
enumerating its elements, regardless of their location. 
Each \texttt{BDBag} has a \texttt{data/} directory containing 
(meta)data files, along with a checksum for each file.
A \texttt{minid} for each of these \texttt{BDBags} provides 
a lightweight persistent identifier 
for unambiguously identifying the dataset regardless 
of their location. Computing a checksum of \texttt{BDBag} contents 
allows others to validate that they have the correct dataset, and that there
is no loss of data. In short, these tools enable us to package and describe our 
datasets in a common manner, and to refer unambiguously to 
the datasets, thereby enabling efficient management and exchange 
of data.

\begin{enumerate}[nosep]
\item \texttt{BDBag} for training set~\cite{BDBag_training} and its associated \texttt{minid:olgmRyIu8Am7}
\item \texttt{BDBag} for validation set~\cite{BDBag_validation} with its associated \texttt{minid:16RmizZ1miAau}
\end{enumerate}

\vspace{3mm}

\noindent In addition to the FAIR properties listed above, 
we have ensured that our datasets meet the detailed 
FAIR guidelines described in Ref.~\cite{fair_hbb_dataset}.

\subsection*{FAIR AI models} The understanding of FAIR 
principles in the creation and sharing of AI models is an active 
area of research. Here we contribute to this effort by 
introducing a set of measurable FAIR principles for AI models, 
with an emphasis on practical applications. We propose that 
all these principles are quantified as Pass or Fail. 
We have constructed the definitions of FAIR principles for 
AI models taking 
into account the diverse needs and level of expertise of AI practitioners. In practice, we have considered three 
types of AI models, which have the same AI architecture, 
as described in this GitHub repository~\cite{zhengchun_repo}, 
and which are trained with the same  Training\_Set~\cite{braggnn-training-set}. However, as shown in Figure~\ref{fig:fair_ai} 
these AI models have distinct features: (i) a baseline AI model 
was trained using \texttt{PyTorch} on GPUs in the ThetaGPU 
supercomputer at ALCF; 
(ii) the fully trained AI model in (i) was optimized with 
\texttt{NVIDIA TensorRT}, a software development kit (SDK) that 
enables low latency and high 
throughput AI-inference, in the ThetaGPU 
supercomputer at ALCF; and (iii) the AI model 
in (i) was trained using the SambaNova DataScale\textsuperscript{\textregistered} system at the 
ALCF AI-Testbed. These cases 
exemplify different needs and levels of expertise in the 
development of AI tools. However, we show below that irrespective 
of the skill set of AI practitioners, hardware used, 
and target application, all these fully trained 
AI models deliver consistent results. In what follows, 
an AI model refers to a fully trained AI model. 

\vspace{3mm}

\noindent \underline{\sffamily{Findable}} 
\noindent \textbf{\textit{Proposition}} An AI model is 
findable when a DOI may direct a human or 
machine to a digital resource that contains all the required 
information to define uniquely the AI model, i.e., 
descriptive and rich \textbf{AI model metadata} that provides 
the title of 
the model, authors, DOI, year of publication, 
free text description, information 
about the input and output data type 
and shape, dependencies (\texttt{TensorFlow}, \texttt{PyTorch}, 
\texttt{Conda}, etc.) and their versions used to create 
and containerize the AI model. The published AI model 
should also include instructions 
to run it, a minimal test set to 
evaluate its performance, and the actual fully trained 
AI model in a user-/machine-friendly 
format, such as a Jupyter notebook or a container.

\noindent \textsc{This work} We have published three AI models in DLHub, and assigned DOIs to each of them: 
(i)  a traditional \texttt{PyTorch} model; 
(ii) an \texttt{NVIDIA TensorRT} version of the 
traditional \texttt{PyTorch} model; and 
(iii) a model trained on the SambaNova DataScale\textsuperscript{\textregistered} system:
\vspace{2mm}
\begin{enumerate}[nosep]
    \item \texttt{PyTorch Model}~\cite{pt_bnn_model}
    \item \texttt{TensorRT Model}~\cite{trt_bnn_model}
    \item SambaNova Model~\cite{sn_trt_model}
\end{enumerate}
\vspace{3mm}

\noindent Each of these 
AI models includes rich and descriptive metadata following 
the DataCite metadata standard, and is available as JSON 
formatted responses through a REST API or Python SDK. Furthermore, DLHub 
provides an interface that enables users to seamlessly 
run AI models following step-by-step examples, 
and explore the AI model's metadata in detail.

\vspace{3mm}
\noindent \underline{\sffamily{Accessible}} \textbf{\textit{Proposition}} 
An AI model is accessible when it is discoverable by a human 
or machine, and it may be downloaded (to further develop it, 
retrain it, optimize it for accelerated inference, etc.) 
or directly invoked to conduct AI inference. The AI model's 
metadata should also be discoverable even in the absence of the 
AI model. 

\vspace{0.5ex}

\noindent \textsc{This work} Our AI models are freely accessible 
through their DOIs provided by DLHub. 
The AI models may be freely downloaded or invoked over the 
network for AI inference. 
To do the latter, we have deployed 
funcX endpoints 
at the ThetaGPU supercomputer, 
which may be used to invoke the AI models and 
access the datasets that we have published at 
DLHub and the MDF, respectively. 
Users may interact with published AI models 
in DLHub by submitting 
HTTP requests to a REST API. In essence, the DLHub SDK 
contains a client that provides a Python API to these 
requests and hides the 
tedious operations involved in making an HTTP call from Python. 
The schema used 
by DLHub is such that 
the AI model's metadata provides explicit information 
about the AI model's identifier and its type (in 
this case a DOI). Thus, this schema ensures that the 
AI model's metadata remains discoverable, even in the 
absence of the AI model.
\vspace{3mm}

\noindent \underline{\sffamily{Interoperable}} \textbf{\textit{Proposition}} 
An AI model is interoperable when it may be readily used  
by machines to conduct AI-driven inference across 
disparate hardware architectures. This property may be realized by 
containerizing the model and providing 
infrastructure to enable AI models to process data in 
disparate hardware architectures. 
Furthermore, 
the AI model's metadata should use a formal, accessible, 
and broadly used format, such as JSON or HTML. 

\vspace{0.5ex}

\noindent \textsc{This work} We have quantified this property 
by evaluating the performance of our AI models across disparate 
hardware architectures. For instance, we have run our 
three AI models using RDUs, CPUs, and GPUs available 
at ALCF. Furthermore, as we describe below, 
we published these 
models in DLHub 
with metadata available as both JSON and HTML. 
Furthermore, the 
AI models' metadata in 
DLHub follows 
the vocabulary of the DataCite metadata standard.
\vspace{3mm}

\vspace{0.5ex}

\noindent \underline{\sffamily{Reusable}} \textbf{\textit{Proposition}} 
An AI model is reusable when it may be used by 
humans or machines to reproduce its putative capabilities 
for AI-driven analyses, and when 
it contains quantifiable metrics that inform users whether 
it may be used to process datasets that differ from those 
originally 
used to create it. This property can be achieved by providing 
information about the AI model's required input and output data 
types and shapes; 
examples that show how to invoke the model; 
and a control dataset and uncertainty quantification 
metrics that indicate the realm of usability of 
the model. 
These reusability metrics may also be used 
to identify when a model is no longer trustworthy, 
in which case active learning, transfer learning, or 
related methods may be needed to fine tune 
the model so it may provide trustworthy predictions. 
The AI model's metadata should also provide detailed 
provenance about the AI model, i.e., authors, dependencies 
and their versions used to create and containerize 
the model, datasets used to train the model, 
year of publication, and a brief description 
of the model.  

\vspace{0.5ex}

\noindent \textsc{This work} Our models published in DLHub 
include examples that describe the input data type 
and shape, the output data type and shape, and a 
sample dataset to quantify their performance. We have also 
provided domain-informed metrics to ascertain when 
the predictions of our AI models are trustworthy. 
DLHub SDK provides 
step-by-step examples that show how to run the models 
and explore the AI model's metadata, which provides 
detailed provenance of the AI model.
In this study we 
use the L2 norm or Euclidean distance, a well-known 
metric for quantifying
the performance of AI models for regression. 
\vspace{3mm}

\paragraph{Expected outcome of proposed FAIR principles 
for AI models.} In the same vein as FAIR principles for 
scientific data aim to automate data management, the FAIR 
principles for AI models we proposed above aim to 
maximize the impact of AI tools and methodologies, and 
to facilitate their adoption and further development 
with a view to eventually realize
autonomous AI-driven scientific discovery. To realize that 
goal it is essential to link FAIR and AI-ready datasets 
with FAIR AI models within a flexible and smart computing 
fabric that streamlines the creation and use of 
AI models in scientific discovery.
The domain-agnostic computing framework we described above 
represents a step in that direction, since it 
brings together scientific data infrastructure (DLHub \& funcX \& Globus), modern computing 
environments (ALCF and ALCF AI-Testbed), and leverage FAIR \& AI-ready 
datasets (MDF).

This ready-to-use framework addresses common challenges 
in the adoption and use of AI methodologies, namely,  
it provides examples that illustrate how to use AI models, 
and their expected input and output data; it 
enables users to readily use, download, or 
further develop AI models. We have created this 
framework because, as AI practitioners, 
we are acutely aware that a common roadblock to using 
existing AI models is that of deploying a model on a computing 
platform but finding that, for example, library incompatibilities
slow down progress and lead to duplication of efforts, 
or discourage researchers from investing limited time and resources 
in the adoption of AI methodologies. Our proposed 
framework addresses these limitations, and demonstrates how to 
combine computing platforms, container technologies, and 
tools to accelerate AI inference. 
Our proposed framework also highlights the importance of 
including uncertainty quantification metrics 
in AI models that inform users on the 
reliability and realm of applicability of AI predictions.

\section*{Methods}

\subsection*{Dataset Description}
\noindent The sample dataset we used to train and 
evaluate BraggNN~\cite{BraggNN-IUCrJ} was collected by using 
an undeformed bi-crystal Gold sample~\cite{ShadeGold} with 
1440 frames (0.25$^\circ$ steps over 360$^\circ$) 
totaling \num{69347} valid Bragg peaks. Each image is 
\(11\times11\) pixels. The spatial resolution of the data 
is such that each pixel is \(200\mu\textrm{m}\).
We used \(80\%\) of 
these peaks (\num{55478}) as our training set, \num{6000} 
peaks ($\sim$9\%) as our validation set for early 
stopping~\cite{goodfellow2016deep}, and the remaining 
\num{7869} peaks ($\sim$11\%) as a test set. We also created a smaller validation dataset consisting of 13799 samples taken from the training dataset, which we used to compute and report relevant metrics. 

\subsection*{AI Models Description}
We present three AI models, namely, (i) a traditional \texttt{PyTorch} 
model; (ii) an NVIDIA \texttt{TensorRT} engine that optimizes  
our traditional \texttt{PyTorch} 
for accelerated inference; and (iii) a model trained with 
the SambaNova DataScale\textsuperscript{\textregistered} system 
at the ALCF AI-Testbed.

\vspace{0.5ex}

\noindent \textbf{PyTorch Model}
Our base AI model is implemented with 
the \texttt{PyTorch} framework. We trained
the model for 500 epochs with a mini-batch size of 512, 
and used validation-based early stopping to
avoid overfitting. Training takes around two hours with 
an NVIDIA V100 GPU in the ThetaGPU supercomputer. 
As mentioned above, the input data 
are images \(11\times 11\) pixels. The AI model's output is a 
2D list of Bragg peak positions in terms of pixel 
locations. 

\vspace{0.5ex}

\noindent \textbf{Model Conversion to \texttt{TensorRT}}
We use the \texttt{Open Neural Network Exchange 
(\texttt{ONNX})} converter within \texttt{PyTorch} 
to convert our PyTorch 
BraggNN model into the \texttt{ONNX} format. We then use \texttt{TensorRT} by 
invoking a Singularity container~\cite{singularity_osti} 
that includes 
\texttt{TensorRT}, \texttt{ONNX}, \texttt{PyTorch}, \texttt{PyCUDA} and other common deep learning 
libraries to build a \texttt{TensorRT} engine and save it in a .plan 
format. We used the following parameters when building our 
engine: the maximum amount of memory that can be allocated 
by the engine, set to 32 GB, allowed the \texttt{TensorRT} graph 
optimizer to make opportunistic use of half-precision 
(FP16) computation when feasible; 
input dimensions, including the batch size,  of \((16384, 1, 11, 11)\);
output dimensions of \((16384, 1, 11, 11)\); and a flag 
that serializes the built engine so that the engine 
will not have to be reinitialized in subsequent runs. 
\texttt{TensorRT} applies a series of optimizations to 
the model by running a GPU profiler to find the best GPU 
kernels to use for various neural network computations, 
applying graph optimization techniques to reduce the 
number of nodes and edges in a model such as layer 
fusion, and quantization where appropriate. Finally, 
we create an inference script using \texttt{PyCUDA} that allocates 
memory for the model and data on the GPU and makes the 
appropriate memory copies between the CPU and GPU in order 
to perform accelerated inference with our \texttt{TensorRT} engine.

\vspace{0.5ex}

\noindent \textbf{SambaNova Model}
The SambaNova 
DataScale\textsuperscript{\textregistered} system at the ALCF AI-Testbed 
uses SambaFlow\textsuperscript{TM} software, 
which has been 
integrated with popular open source APIs, such as 
\texttt{PyTorch} and \texttt{TensorFlow}. Leveraging 
these tools, we used SambaFlow to automatically 
extract, optimize, and execute our originally 
\texttt{PyTorch} BraggNN model with 
SambaNova's RDUs~\cite{liu2021bridge}. 
We find that the predictions of our SambaNova BraggNN 
model are consistent with those obtained with 
\texttt{PyTorch} and \texttt{TensorRT} models.

\begin{figure}[h]
\centerline{
\includegraphics[width=\textwidth]{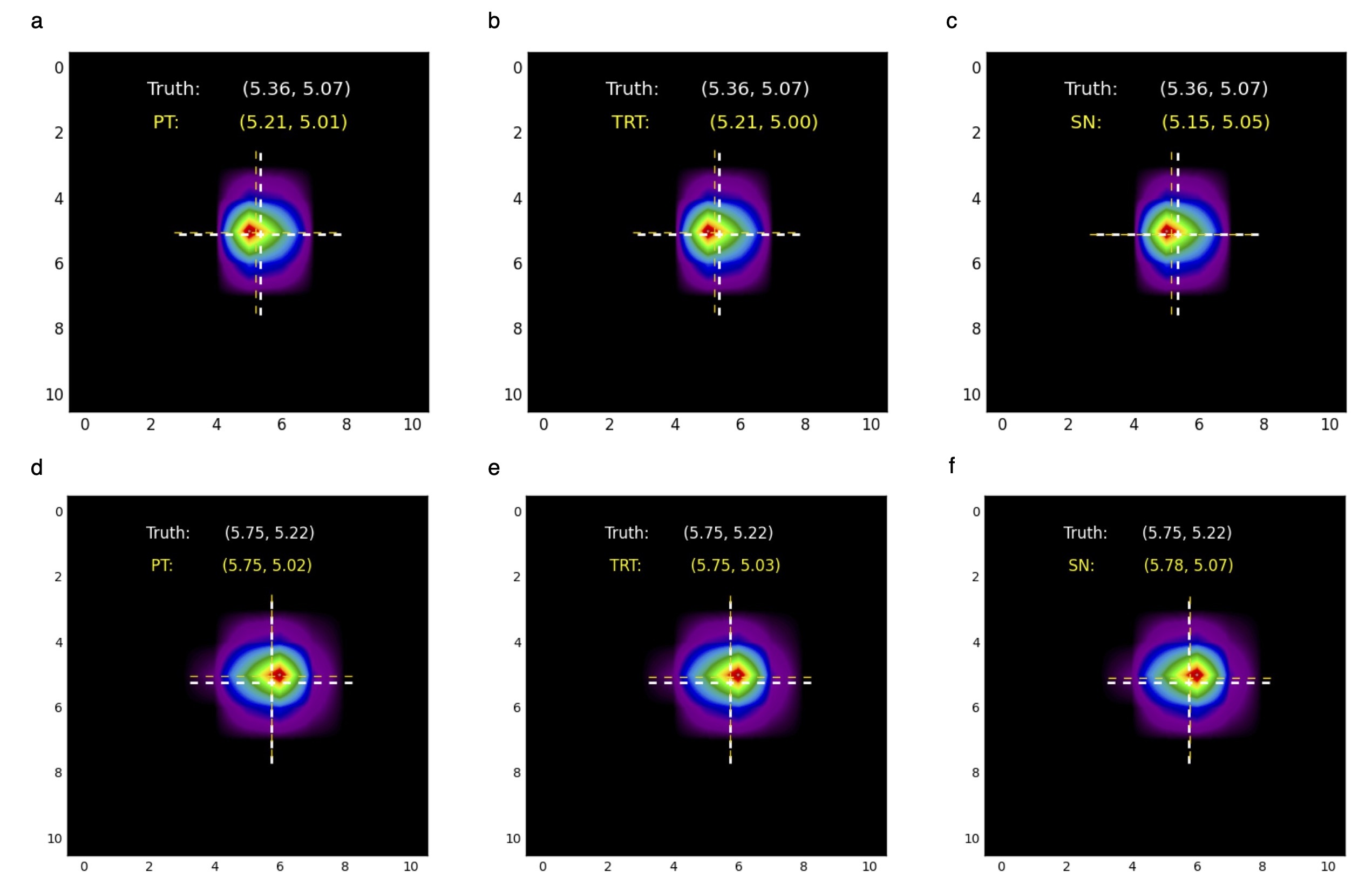} 
}
    \caption{\textbf{Bragg peak reconstruction. Top panels.} Inference results for the identification of Bragg peak locations in an 
    undeformed bi-crystal gold sample. From left to right, we show 
    results for three AI models, namely:  \textbf{(a)} \texttt{PyTorch}  (\texttt{PT}) baseline 
    model; \textbf{(b)} an inference optimized \texttt{TensorRT} (\texttt{TRT}) model; and \textbf{(c)} a model 
    trained with the SambaNova DataScale\textsuperscript{\textregistered} (\texttt{SN}) system. 
    In the panels, \texttt{Truth} 
    stands for the ground truth location of Bragg peaks; 
    \texttt{PT}, \texttt{TRT} and \texttt{SN} represent the 
    predictions of our baseline 
    \texttt{PyTorch}, \texttt{TensorRT} and SambaNova 
    models, respectively. We produced these results by directly 
    running these models in the ThetaGPU supercomputer, and 
    found that 
    \(95\%\) of the predicted peak locations in the test set are 
    within a Euclidean distance of \(0.688\, \textrm{pixels}\) from 
    the actual peak locations. \textbf{FAIR AI Approach. Bottom panels.} AI inference results obtained 
    by combining DLHub, funcX and the ThetaGPU supercomputer. 
    From left to right, we show results for our three AI models:  \textbf{(d)} \texttt{PT}; \textbf{(e)} 
    \texttt{TRT}; and \textbf{(f)} \texttt{SN}, which are hosted at DLHub. 
    funcX manages the entire workflow by invoking AI models, launching workers 
    in ThetaGPU and doing AI inference on a test set. This 
    workflow also includes post-processing scripts to
    quantify the L2 norm that provides a measure for the 
    reliability of our AI-driven regression analysis.} 
    \label{fig:consistent_AI_models}
\end{figure}

\vspace{0.5ex}

\noindent \textbf{Benchmark results in the ThetaGPU supercomputer} 
We used these three models to conduct AI inference in the 
ThetaGPU supercomputer using the validation dataset described above, i.e., 
13799 Bragg peaks. We quantified the consistency of their predictions 
using Euclidean distance between the predicted peak locations 
and ground truth peak locations. We found that for all three models, 
\(95\%\) of the predicted peak locations in our test set are within 
a Euclidean distance of \(0.688\) pixels from the actual 
peak locations. In addition, the average Euclidean error is only $\approx$ \(0.21\) pixels, 
and the standard deviation of the Euclidean error is $\approx$ \(0.22\) pixels. For reference, 
the images used for this study 
are \(11\times11\) pixels, and each pixel is \(200 \mu\mathrm{m}\) 
in size. Therefore, these results show that our three different models 
produce accurate and consistent predictions, even if they are trained 
using different optimization schemes or hardware architectures. 
We show a sample of these results in the top three panels of 
Figure~\ref{fig:consistent_AI_models}.

\noindent \textbf{AI models in DLHub}
AI models published in DLHub are containerized 
by using Docker, and include instructions for 
running the models with a sample 
test set. The models include uncertainty quantification 
metrics to inform users about their expected 
performance and realm of applicability.  
All trained AI models are assigned a DOI, and include 
descriptive metadata including the title, authors, free text 
description, and more (following the DataCite metadata standard), 
input type and shape (e.g., [11,11] image maps), output type and 
shape (e.g., [1,2] list of predicted Bragg peak positions) as 
well as examples for how to invoke each individual model. 
These metadata are available as JSON formatted responses 
through a REST API or Python SDK; or as HTML through a 
searchable web interface. We find that running our AI models via DLHub 
yields inference results that are consistent with those obtained when the models 
are run natively on ThetaGPU.

\begin{figure}[ht]
\centerline{
\includegraphics[width=\textwidth]{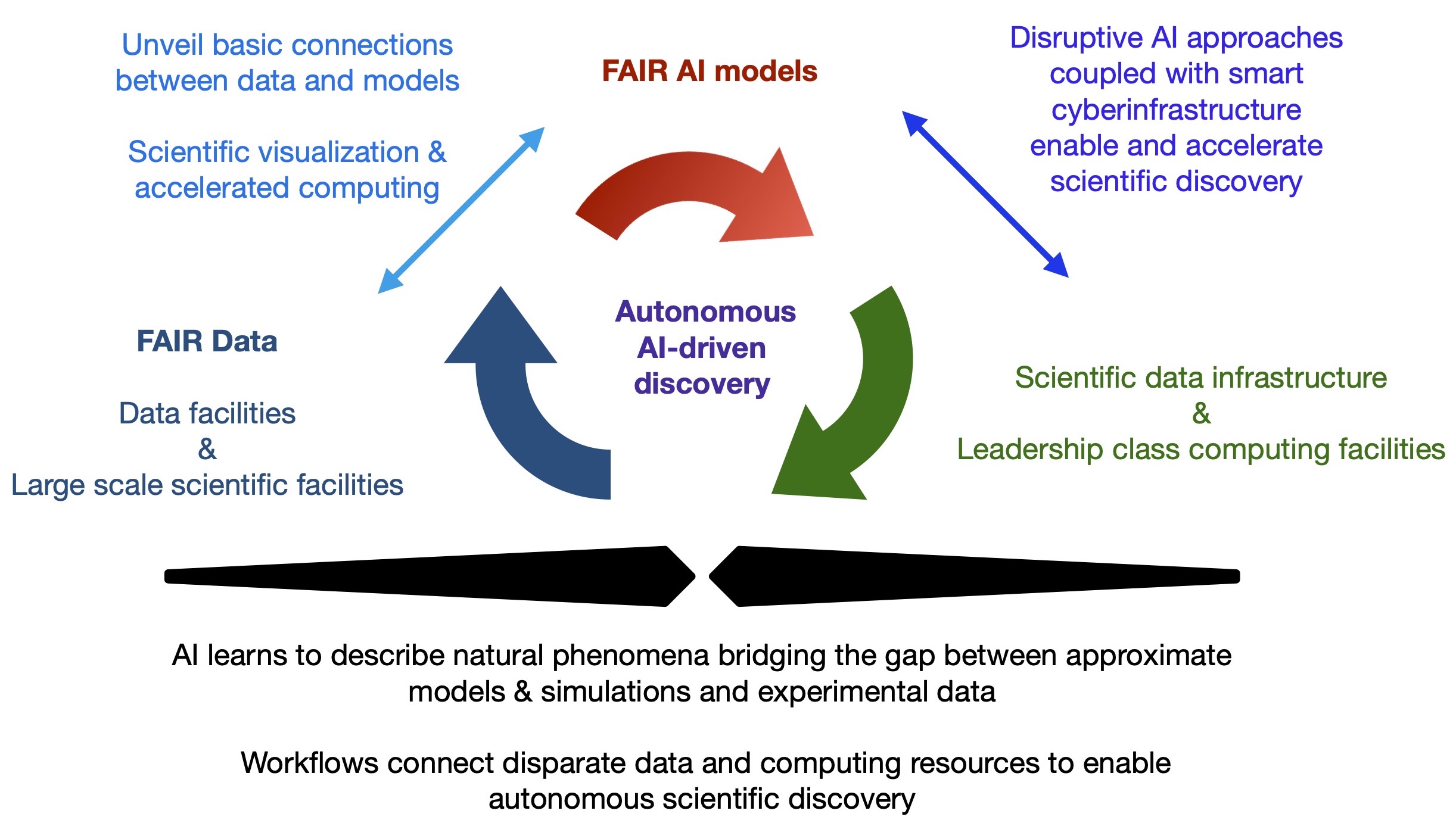}
}
    \caption{\textbf{Autonomous AI-driven discovery} Vision for
    the integration of FAIR \& AI-ready datasets with FAIR 
    AI models and modern computing environments to enable 
    autonomous AI-driven discovery. This approach will 
    also catalyze the development of next-generation AI methods, and the creation of a rigorous approach that identifies foundational connections between data, models and 
    high performance computing.} 
    \label{fig:comp_frame}
\end{figure}

\vspace{0.5ex}

\noindent \textbf{DLHub, funcX, and ThetaGPU}
DLHub is configured to perform on-demand inference in Docker 
containers on a Kubernetes cluster 
hosted at the University of Chicago. 
The DLHub execution model leverages funcX~\cite{chard2020funcx}, a federated 
function as a service (FaaS) platform, that enables fire-and-forget remote execution. In this work, we extended 
that model to use
a funcX endpoint deployed on ThetaGPU and configured to 
dynamically provision resources from ThetaGPU. ThetaGPU is 
an extension of the Theta supercomputer and consists of 
24 NVIDIA DGX A100 nodes. Each DGX A100 node has eight NVIDIA 
A100 Tensor Core GPUs and two AMD Rome CPUs that provide 22 
nodes with 320 GB of GPU memory and two nodes with 640 GB 
of GPU memory. Access to ThetaGPU is currently restricted by policy to authorized ALCF users. 
One approach to enabling broader access in the future could be to configure the funcX 
endpoint to provide access to members of a Globus Group~\cite{chard2016globus}. 
Users could then request access to this group and, following an approval process, be 
granted access to run the models on ThetaGPU.

Finally, ALCF does not support Docker containers as Docker 
requires root privileges. As a result, we first had to transform 
DLHub servable containers into the Apptainer (previously 
Singularity) containers supported by ALCF. 
Apptainer provides an effective security model whereby users 
cannot gain additional privileges on the host system, making 
them suitable for deployment on high performance computing 
resources. After creating Apptainer  containers for each 
AI model, 
we registered each with funcX along with the associated 
DLHub invocation function, enabling on-demand inference of the 
models using ThetaGPU.

\vspace{0.5ex}

\noindent \textbf{Reproducibility of AI models}
The computational framework---DLHub, funcX and 
ALCF---provides a 
ready to use, user friendly solution 
to harness AI models, FAIR datasets, and 
available computing resources to enable 
AI-driven discovery. 
We have tested 
the reliability of this computational framework for AI-driven discovery 
by processing a test set with each AI model, 
finding that results across models are consistent, 
and that these results are the same as those obtained 
by running AI models directly in the ThetaGPU supercomputer.
The uncertainty quantification metrics 
we have included in our AI models also guide researchers 
in the use and interpretation of these AI predictions. 
A sample of these results is shown in the bottom panels of 
Figure~\ref{fig:consistent_AI_models}.

\vspace{0.5ex}

\noindent \textbf{Readiness and usability of computational framework}  We 
asked non-developers of this framework to test it and to provide feedback. They reported 
no issues when they followed step-by-step instructions contained in a Jupyter notebook that 
indicates how to load experimental data, invoke AI models from DLHUb, and then use a funcX 
endpoint at ThetaGPU to do inference and then compute L2 results for uncertainty quantification. 
With this ready-to-use, user-friendly notebook, they were able to reproduce the results we 
report in this article. It is worth pointing out that test users utilized their own ALCF allocation 
to conduct this exercise.

\section*{Discussion}

We have showcased how to FAIRify experimental 
datasets by harnessing the MDF and using 
\texttt{BDBags} and \texttt{minids}. Using these 
FAIR and AI-ready datasets, we described how to 
create and share FAIR AI models using a new set of 
practical definitions that we introduced in this article. 
Throughout this analysis, we have showcased how to harness 
existing data facilities, FAIR tools, modern computing 
environments and scientific data infrastructure to 
create a computational framework that is conducive for 
autonomous AI-driven discovery. To realize that goal, 
our FAIR and AI-ready datasets are published including 
Jupyter notebooks that provide key information regarding 
data type, shape and size, and how these datasets may be 
readily used for the creation and use of AI 
models. Complementing these approaches, our AI models published 
in DLHub include examples that 
illustrate how to use FAIR and AI-ready datasets for AI inference. 
The models also include uncertainty quantification 
metrics to ascertain their validity, reliability, reproducibility and 
statistical robustness.

The FAIR for AI data and models approach that we describe in this 
article focuses on the creation of a ready-to-use, user-friendly 
computational framework in which data, AI, and 
computing are indistinguishable components of the same fabric. 
To realize this vision, we have brought 
together DLHub, 
MDF, funcX, and ALCF. 
Anticipating that this framework will be of interest to 
researchers and AI practitioners who are eager to explore 
and incorporate FAIR best practices 
in their research programs, we release with this 
manuscript all scientific software, data, and AI models 
used in this work. We hope that this framework is harnessed 
and further developed 
by researchers to enable advances in science, engineering 
and industry, as illustrated in Figure~\ref{fig:comp_frame}.

\section*{Data Availability}
Our FAIRified datasets are published at the Materials 
Data facility: Training\_Set~\cite{braggnn-training-set} and Validation\_Set~\cite{braggnn-validation-set}. 
All FAIRified BraggNN datasets are 
released in HDF5 format.

\section*{Code Availability}
\noindent The three AI models introduced in this article, 
along with the Jupyter notebook and scientific software 
needed to reproduce our results have been released through 
DLHub~\cite{pt_bnn_model,trt_bnn_model,sn_trt_model}. Scientific 
software to train these models may be found in an open source GitHub repository~\cite{zhengchun_repo}.
\vspace{3mm}

\noindent Received: 14 July 2022; Accepted: 21 September 2022;\\
Published online: 10 November 2022

\bibliography{references}

\section*{Acknowledgements}
\noindent This work was supported by the FAIR Data 
program and the Braid project of the U.S. Department of Energy, 
Office of Science, Advanced Scientific Computing 
Research, under contract number DE-AC02-06CH11357. 
It used resources of the Argonne 
Leadership Computing Facility, which is a DOE Office of 
Science User Facility supported under Contract 
DE-AC02-06CH11357.  \textbf{Dataset Publication:} This work was performed under financial assistance award 70NANB14H012 from U.S. Department of Commerce, National Institute of Standards and Technology as part of the Center for Hierarchical Material Design (CHiMaD), and by the National Science Foundation under award 1931306 ``Collaborative Research: Framework: Machine Learning Materials Innovation Infrastructure.''
\textbf{Model Development:} R\&D for the creation of AI models 
presented in this article was supported 
by Laboratory Directed Research and Development (LDRD) 
funding from Argonne National Laboratory, 
provided by the Director, Office of Science, of the 
U.S. Department of Energy under 
Contract No. DE-AC02-06CH11357. We also thank SambaNova Systems, 
Inc., for engineering support to make our BraggNN AI models 
work efficiently on their system.
\textbf{Model Publication:} Development of DLHub has been supported by LDRD 
funding from Argonne National Laboratory, provided by the Director, Office of Science, 
of the U.S. Department of Energy under Contract No. DE-AC02-06CH11357. This 
work was also supported by the National Science Foundation under NSF 
Award Number: 2209892 ``Frameworks: Garden: A FAIR Framework for Publishing 
and Applying AI Models for Translational Research in Science, Engineering, Education, and Industry''
\textbf{Computational 
framework testing:} We thank Roland Haas for 
independently testing, and providing feedback, on the 
readiness and usability of the methodology and 
computational framework proposed in this article 
to quantify the FAIRness of AI models.

\section*{Author contributions statement}
E.A.H. conceived and led this work. N.R. and Z.L. developed 
\texttt{PyTorch} BraggNN models. N.R. ported \texttt{PyTorch} models into 
\texttt{TensorRT} engines. P.C. containerized both \texttt{PyTorch} and 
\texttt{TensorRT} models and ran computing tests on the 
ThetaGPU supercomputer. Z.L. developed a BraggNN 
SambaNova model and ran tests on that system. 
N.R. and B.B. curated and published FAIR and AI-ready datasets 
in the Materials Data Facility. N.R. produced \texttt{BDBags} and 
\texttt{minids} for these datasets. K.C. provided guidance on 
the use of \texttt{BDBags} and 
\texttt{minids} and funcX.
B.B., R.C., A.S. and K.J.S. 
deployed all AI models in DLHub 
and conducted 
tests connecting DLHub to ThetaGPU by using funcX. 
I.F. guided the creation of FAIR datasets and 
models. All 
authors contributed ideas and participated in the 
writing and review of this manuscript.

\section*{Competing interests} 
The authors declare no competing financial and/or non-financial interests in relation to the work described in this manuscript.

\end{document}